\documentclass[10pt,twocolumn,letterpaper]{article}

\usepackage[pagenumbers]{cvpr} 
\usepackage{lipsum}
\usepackage{amsmath}
\usepackage{booktabs}
\usepackage{multirow}
\usepackage{makecell}
\usepackage{graphicx}
\newcommand{\name}{KLIP}

\newcommand{\block}{block}
\newcommand{\Block}{Block}
\newcommand{\blocks}{blocks}
\newcommand{\dkl}[2]{D_{KL}(#1 \Vert #2)}

\newcommand{\idset}{\textit{ID set}}
\newcommand{\oodset}{\textit{OOD set}}
\newcommand{\calibrationset}{\textit{Tuning set}}

\newcommand{\songAlgOneStar}{0.999}
\newcommand{\songAlgTwoStar}{0.993}
\newcommand{\songAlgOneStarimg}{0.978}
\newcommand{\songAlgTwoStarimg}{0.999}
\newcommand{\songAlgOneTumor}{0.505}
\newcommand{\songAlgTwoTumor}{0.504}
\newcommand{\songAlgOneTumorimg}{0.441}
\newcommand{\songAlgTwoTumorimg}{0.592}

\newcommand{\songNLLstar}{0.586}
\newcommand{\songDiffpathstar}{0.688}
\newcommand{\songDKLstar}{0.541}
\newcommand{\songoursstar}{0.855}
\newcommand{\songDKLstarimg}{0.837}
\newcommand{\songoursstarimg}{0.912}
\newcommand{\PaDISDKLstarimg}{0.841}
\newcommand{\PaDISoursstarimg}{0.889}

\newcommand{\songNLLtumor}{0.535}
\newcommand{\songDiffpathtumor}{0.368}
\newcommand{\songDKLtumor}{0.602}
\newcommand{\songourstumor}{0.776}
\newcommand{\songDKLtumorimg}{0.856}
\newcommand{\songourstumorimg}{0.878}
\newcommand{\PaDISDKLtumorimg}{0.672}
\newcommand{\PaDISourstumorimg}{0.732}

\newcommand{\CelebADKLsyntheticimg}{0.675}
\newcommand{\CelebADKLcollectedimg}{0.482}
\newcommand{\CelebAourssyntheticimg}{0.867}
\newcommand{\CelebAourscollectedimg}{0.772}

\definecolor{cvprblue}{rgb}{0.21,0.49,0.74}
\usepackage[pagebackref,breaklinks,colorlinks,allcolors=cvprblue]{hyperref}

\def\paperID{ } 
\def\confName{CVPR}

\title{KLIP: localized distribution shift detection via KL-divergence\\with diffusion priors in Inverse Problems}

\author{Alireza Kheirandish*
\hspace{1.5cm}
Jihoon Hong*
\hspace{1.5cm}
Sara Fridovich-Keil\\
Georgia Institute of Technology\\
School of Electrical and Computer Engineering\\
{\tt\small \{akheirandish3, jhong392, sfk\}@gatech.edu}
}

\begin{document}

\twocolumn[{%
  \renewcommand\twocolumn[1][]{#1}%
  \maketitle
}]

\footnotetext{Equal contribution. Order of names was decided by a coin toss.}

\begin{abstract}
Diffusion models have shown promising performance as data-driven priors for computational imaging, as well as some capacity to detect out-of-distribution (OOD) images.
However, existing approaches to OOD detection often require some knowledge of the shifted distribution, fail to detect subtle or localized distribution shifts, and operate on full images, rather than the indirect measurements available in inverse problems.
We propose an OOD detection metric based on the Kullback-Leibler divergence between the diffusion prior and the posterior distribution, that (i) does not require any calibration data or knowledge of the shifted distribution, and
(ii) can detect whole images as OOD as well as localize OOD patches within an image. 
Experimentally, we show that this metric can detect subtle yet semantically meaningful distribution shifts, such as the shift from healthy liver CT scans to those with tumors, and generalizes across different types of diffusion models, datasets, and inverse problems. Our code can be found at \url{https://github.com/voilalab/KLIP}.

\end{abstract}    

\begin{figure}[t]
  \centering
  \includegraphics[width=\linewidth]{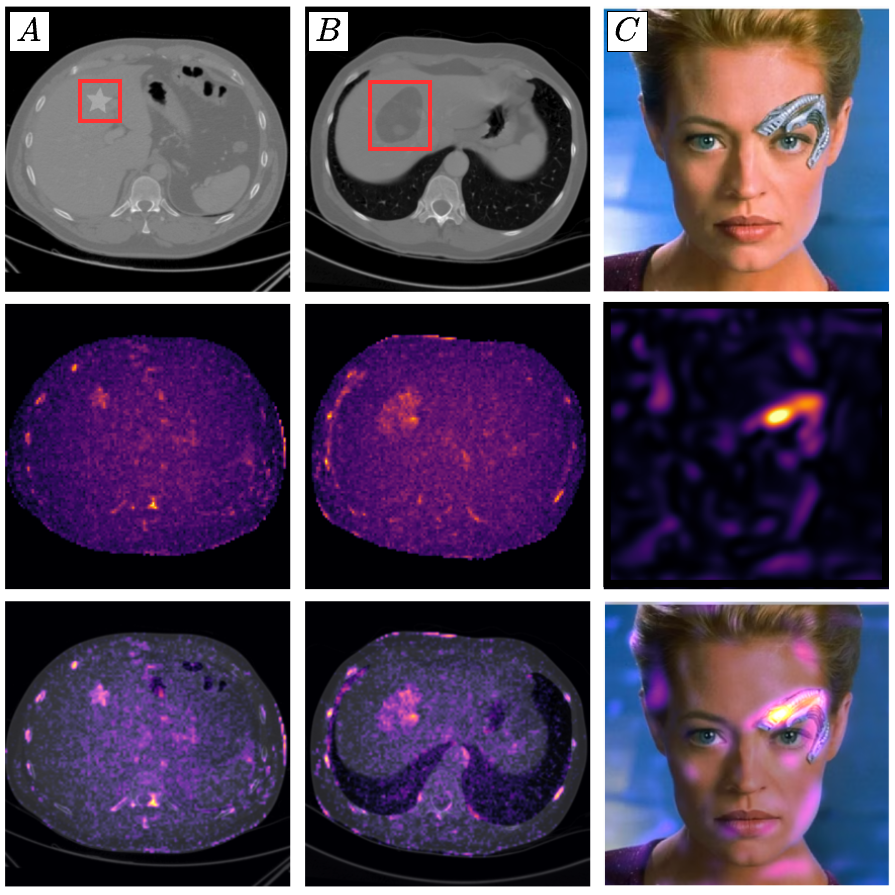}
  \caption{\name{} (\emph{middle row}) is able to identify the precise locations of local distribution shifts from diffusion model samples acquired through posterior sampling in an inverse problem (here, underdetermined CT and Gaussian deblurring). The diffusion model is trained on only ID images (healthy CT scans and celebrity faces), and \name{} has no access to other OOD examples.
    }
  \vspace{-3mm}
  \label{fig:teaser}
\end{figure}

\vspace{-1em}
\section{Introduction}
\label{sec:intro}

Distribution shift detection is a foundational problem in machine learning and computational imaging, especially when models are to be deployed in an open world. 
In many applications, the distribution shifts that are most important to catch are also some of the most difficult to detect, often involving subtle or localized changes in an image. 
In medical imaging, for example, out-of-distribution (OOD) features may consist of small lesions, tumors, or tears that are essential to the diagnostic value of the image. 
We propose to use diffusion models trained on in-distribution (ID) images to simultaneously perform underdetermined image reconstruction and detection of spatially localized OOD features.

Many approaches have been proposed for OOD detection using diffusion models. 
The most statistically rigorous OOD detection metrics are often based on conformal prediction \cite{angelopoulos2024conformal, angelopoulos2022imagetoimage, kutiel2023conformal}, but these require access to a calibration dataset of OOD images, which is impossible to acquire for the long tail of rare abnormalities that appear in open-world images. 
Other approaches to OOD detection are more heuristic, but still require an OOD calibration dataset \cite{ming2022poem}, require training multiple diffusion models \cite{shoushtari2025unsupervised}, operate on full images rather than indirect measurements available in computational imaging \cite{diffpath}, or focus on distinguishing OOD images with global rather than local OOD features \cite{shoushtari2025eigenscore}.
We instead propose an OOD detection metric that (i) does not require any knowledge or samples of OOD images, (ii) can be applied to indirect measurements of an image that are available at inference time in an inverse problem, and (iii) can detect small and localized, yet semantically meaningful, distribution shifts.

Our core idea is that the Kullback-Leibler (KL) divergence between the prior distribution of the image, and the posterior distribution of the image given the measurements, should be larger for OOD images and larger for OOD regions within an OOD image. This KL divergence can be estimated using only the update steps generated throughout diffusion sampling conditioned on a set of measurements, requiring no OOD calibration data. 
Using the knowledge that diffusion models generate different components of an image at different stages of the sampling process \cite{chen2024exploring, rissanen2023generative}, we further refine this KL divergence metric by estimating its restriction over specific time windows of the sampling trajectory, and over spatially localized \block{}s in the image.

We evaluate our proposed KL divergence based OOD detection metric, \name{}, on a range of image reconstruction tasks with localized OOD features.
Our primary results focus on sparse view X-ray computed tomography (CT), where the diffusion prior is trained over a dataset of healthy scans. We demonstrate dataset-level and image-level OOD detection of both synthetic stars and realistic synthetic liver tumors \cite{synthetic_tumor}, across two different diffusion model backbones, one trained on whole images \cite{song2021solving} and another trained on image patches \cite{padis}.
We also demonstrate \name{}'s generalization potential via image-level detection of anomalous character facial features in a Gaussian de-blurring task, using a diffusion prior trained on celebrity faces \cite{diffusers}.
As shown in \Cref{fig:teaser}, \name{} can locate the positions of (a) a synthetic star artifact and (b) tumor within a liver CT scan, and (c) the implant around the left eye of Seven of Nine, a character from \emph{Star Trek: Voyager}.
To summarize, we:
\begin{itemize}
    \item Introduce a near-OOD detection metric, KLIP, based on the timestep and \block{} restricted KL divergence between prior and posterior distributions in an inverse problem. We find that these timestep and \block{} restrictions offer a rich window into the sampling process under distribution shift.
    KLIP is calibration-free and can be calculated using a single diffusion model trained on ID images. 
    \item Demonstrate that KLIP can detect dataset-level and image-level distribution shift, when the OOD features are subtle and spatially localized. KLIP generalizes across different inlier distributions (CT scans and human faces), different localized OOD features (synthetic stars, realistic liver tumors, scars, and makeup), and different diffusion model architectures (whole-image and patch-based).
\end{itemize}

\section{Related Works}
\label{sec:related_works}

Many methods have been proposed to leverage diffusion models for distribution shift detection. 

\vspace{-1em}
\paragraph{Calibration-based methods.}
Many approaches to OOD detection, with and without diffusion models, rely on a calibration dataset of OOD images with which to estimate an ID--OOD decision boundary \cite{ming2022poem} or train a separate OOD diffusion model \cite{shoushtari2025unsupervised}. These methods include conformal prediction \cite{angelopoulos2024conformal, horwitz2022conffusion, teneggi2023trust}, which can provide statistically rigorous confidence intervals around OOD image reconstructions. They also include methods that quantify uncertainty at the pixel or region level within an image \cite{angelopoulos2022imagetoimage, kutiel2023conformal}, and adaptive schemes to collect additional measurements to reduce uncertainty \cite{Wen2024TaskDrivenUQ, huang2025cuqds}.
While these methods can work well when both ID and OOD data is available, they are not applicable to distribution shift detection when it first arises, or to detecting the long tail of rare OOD images that exist in an open world.

\vspace{-1em}
\paragraph{Uncertainty quantification methods.}
Many approaches have been proposed for uncertainty quantification, but are not specifically focused on OOD detection or uncertainty arising from distribution shifts. 
For example, Monte Carlo methods have been proposed to estimate uncertainty due to limited measurements \cite{Riis2023CUQIpyIC, Adler2018DeepBI, Adler2019DeepPS}, and ensemble methods have been proposed to quantify uncertainty over the weights of the learned prior \cite{lakshminarayanan2017simple, Jiang2022AnEA}. 
Some approaches to uncertainty quantification avoid the need for a calibration dataset by simulating calibration data with a physical model \cite{Tachella2023EquivariantBF}, especially in inverse problems with strong physical constraints such as partial differential equations \cite{Wu2024UncertaintyQF,Zong2023RandomizedPM,Riis2023CUQIpyIC}.
Our focus is instead on calibration-free detection of local OOD features during image reconstruction, in generic linear inverse problems with diffusion priors.
Existing approaches to calibration-free detection of localized OOD features have focused on detecting anomalies in otherwise highly consistent images, such as those for quality control in manufacturing \cite{hermann2022fast}, rather than in more diverse ID images.

\begin{figure*}[t]
  \centering
    \includegraphics[width=1.0\textwidth]{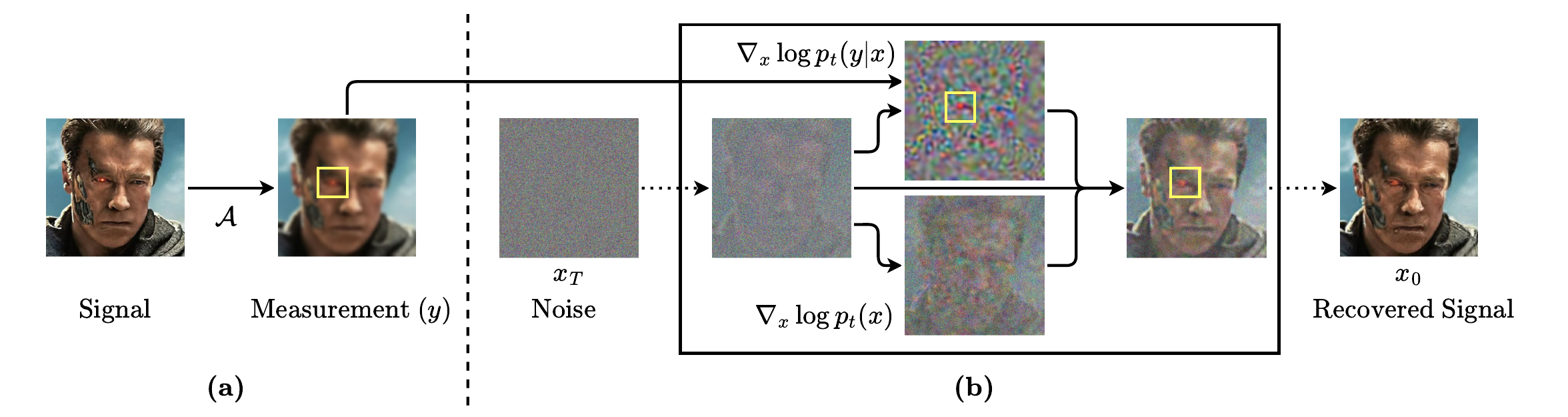}
    \vspace{-5mm}
    \caption{\textbf{Solving inverse problems using posterior sampling with a diffusion model}, for a Gaussian de-blurring example. (a) Forward model of gaussian blur. (b) Posterior sampling using a diffusion model to recover the signal from the measurement $y$ following \Cref{eq:posterior_sampling}. We can see from the yellow boxes highlighting the Terminator's red eye that the score of the likelihood guides sampling to be consistent with $y$, especially where there is a locally OOD feature.}
    \vspace{-3mm}
   \label{fig:preliminary}
\end{figure*}

\vspace{-1em}
\paragraph{OOD detection with generative priors.}
Methods have also been proposed within our focus area of calibration-free OOD detection with generative (specifically diffusion-based) priors, though these generally do not allow for detection of localized OOD features in images that are otherwise in-distribution \cite{diffpath, shoushtari2025eigenscore, vasiliuk2023limitations}. 
A common assumption of OOD detection methods is that OOD samples correspond to regions of low density under the prior distribution \cite{bishop1993novelty}; \cite{nalisnick2019deep} showed that this is not the case for generative models, and \cite{nalisnick2019detecting} proposed an alternative OOD detection test based on the notion of a distribution's \emph{typical set}. While this approach can improve dataset-level OOD detection, it has not been evaluated for capacity to detect the sort of highly localized OOD features that are our focus.

\section{Preliminaries}
\label{sec:preliminaries}

We begin by introducing the diffusion models and posterior sampling methods that are the context for KLIP.

\paragraph{Score-Based Diffusion Models.}
To sample from a data distribution $p(x)$, which is often high dimensional and multi-modal, score-based diffusion models \cite{song2020score, chan2024tutorial} first define two processes: forward and backward.
The forward process is described by a stochastic differential equation (SDE) defined over time $t\in[0,T]$:
\begin{equation}
\label{eq:forward_sde}
dx=f(x,t)dt+g(t)dw ,
\end{equation}
that we call the forward SDE.
Here $f(x,t)$ is the drift coefficient, $g(t)$ is the diffusion coefficient, and $w$ is a standard Brownian motion.
We denote the marginal distribution of $x$ at time $t$ as $p_t(x)$, and $p_0(x)$ is by definition equal to the data distribution $p(x)$.
The coefficients of the SDE are determined such that $p_T(x)$ approximates a tractable distribution that we can easily sample from, typically $\mathcal{N}(0,\sigma_T^2I)$.

The backward process is described by another SDE defined from $t=T$ to $t=0$, also called the reverse SDE of \Cref{eq:forward_sde}, as its marginal distribution at each time coincides with $p_t(x)$.
It takes the form:
\begin{equation}
\label{eq:backward_sde}
dx=\bigl(f(x,t)-g(t)^2\nabla_x\log{p_t(x)}\bigr)dt+g(t)d\overline{w} ,
\end{equation}
where $f(x,t)$ and $g(t)$ are precisely the coefficients of \Cref{eq:forward_sde}, $\nabla_x\log{p_t(x)}$ is the score of the marginal distribution $p_t(x)$ of the forward SDE at time $t$, and $\overline{w}$ is the reverse-time standard Brownian motion.

With the forward and backward processes defined as above, we can easily sample from the data distribution: first sample $x_T$ from $p_T(x)$, and simulate the reverse SDE in \Cref{eq:backward_sde} until we reach $x_0\sim p_0(x)=p(x)$.
The sampling can be performed in various ways, including Euler-Maruyama discretization of the reverse SDE:
\begin{equation}
\label{eq:backward_sampling}
\Delta x_t=\bigl(f(x,t)-g(t)^2s_\theta(x_t;t)\bigr)\Delta t+g(t)z, 
\end{equation}
where $z\sim \mathcal{N}(0,I)$.
Here, since $\nabla_x\log{p_t(x)}$ is not tractable, a diffusion model $s_\theta(x;t)$ trained with a score matching loss is used to approximate it during sampling.

\begin{figure*}[t]
  \centering
    \includegraphics[width=1.0\textwidth]{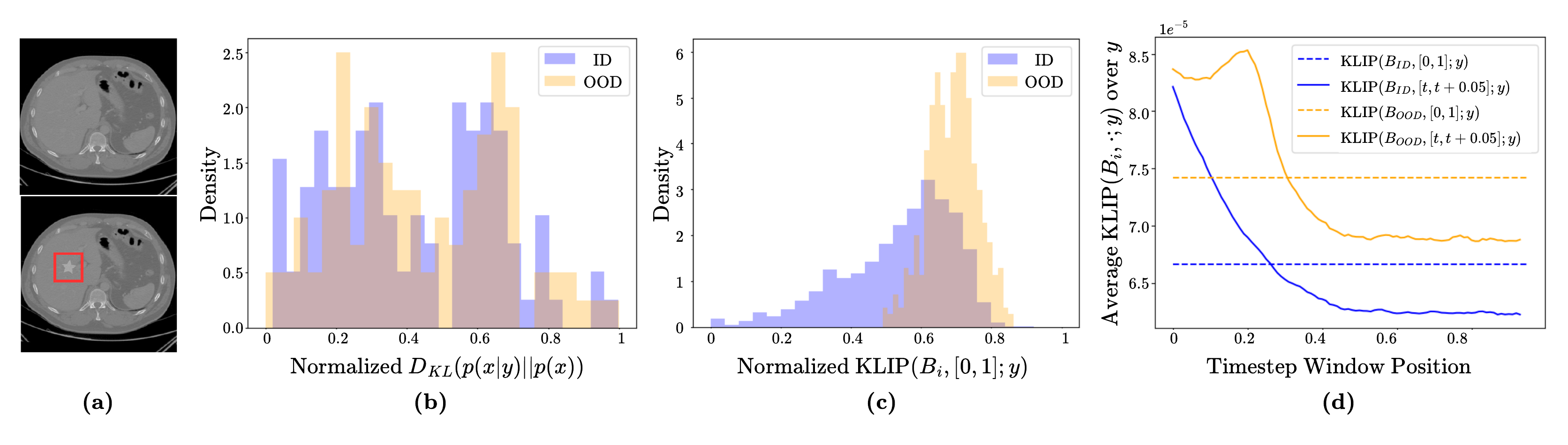}
    \vspace{-7mm}
    \caption{\textbf{Motivation for block and timestep restriction:} 
    (a) An example ID image of a healthy liver CT scan (\emph{top}) and corresponding OOD image with a localized synthetic star artifact (\emph{bottom}). 
    (b) Histogram of the prior--posterior KL divergence (\Cref{eq:dkl_posterior}) for ID images and OOD images, normalized to $[0, 1]$. The ID and OOD images are not well separated. 
    (c) Histogram of the \block{} restricted KL divergence over the same dataset, normalized to $[0, 1]$. OOD \blocks{} are those that contain the synthetic star artifact. These distributions are much better separated, though there is still some overlap. 
    (d) Average of \block{} restricted KL divergence over the entire sampling trajectory (dotted lines) and over a sliding timestep window 
    (solid line). Further restricting the \block{}wise KL divergence to focus on a specific time window can enhance separation between ID and OOD \blocks{}.}
    \vspace{-3mm}
   \label{fig:method_motivation}
\end{figure*}

\paragraph{Inverse Problems.}

Diffusion models are widely used as generative priors to solve inverse problems \cite{dps, song2021solving, dpnp, wang2024dmplug, wu2024principled} of the following standard form:
\begin{equation}
\label{eq:IPForward}
y=\mathcal{A}(x)+ \epsilon,
\end{equation}
where $x$ is an image to be reconstructed, $\mathcal{A}$ models the linear \cite{dps, ddrm} or nonlinear \cite{dpnp, wang2024dmplug, wu2024principled} forward process of collecting measurements $y$, and $\epsilon$ is measurement noise.
These works aim to sample from the posterior distribution $p(x|y)$, using an unconditional diffusion model as well as a measurement $y$ of the ground truth image $x$ acquired through the measurement process in \Cref{eq:IPForward}.
Since the diffusion model is trained to sample from the prior distribution $p(x)$, the sampling process is modified to ensure consistency with the measurements $y$ by replacing the score of the prior distribution $p(x)$ with the score of the posterior $p(x|y)$:
\begin{equation}
\label{eq:posterior_score}
\nabla_x\log{p_t(x|y)}=\nabla_x\log{p_t(x)}+\nabla_x\log{p_t(y|x)} .
\end{equation}
Here we use Bayes' rule to decompose the 
posterior score $\nabla_x\log{p_t(x|y)}$ 
into the sum of the prior score $\nabla_x\log{p_t(x)}$ and the likelihood score $\nabla_x\log{p_t(y|x)}$, which encodes consistency with the measurements $y$.

While the prior score can be estimated using a diffusion model trained on $p(x)$, the likelihood score is typically intractable.
Therefore, previous works 
such as Diffusion Posterior Sampling (DPS) \cite{dps} and Song et al. \cite{song2021solving} 
have proposed approximations or proxies of the likelihood score with a function $s_l(x_t,y;t)$, which we denote with the subscript $l$ for likelihood, and augment the unconditional sampling process as:
\begin{equation}
\label{eq:posterior_sampling}
\Delta x_t=\Bigl(f(x,t)-g(t)^2\bigl(s_\theta(x_t;t) + s_l(x_t,y;t)\bigr)\Bigr)\Delta t+g(t)z.
\end{equation}
DPS approximates the likelihood score directly under assumptions on the measurement noise $\epsilon$, and rigorously proved bounds on the approximation error.
Meanwhile, Song et al. \cite{song2021solving} proposed to insert an additional proximal optimization step at each time $t$ during sampling, and showed experimentally that adding this step is sufficient to sample from $p(x|y)$ instead of $p(x)$ for a limited class of linear forward models.
Therefore, we consider the updates from the optimization step as a proxy for the difference between \Cref{eq:backward_sampling} and \Cref{eq:posterior_sampling}, which is the likelihood score scaled by $g(t)^2$.
\section{Methods}
\label{sec:methods}

\paragraph{Prior--Posterior KL Divergence.}

We posit that the KL divergence between the prior distribution $p(x)$ and the posterior distribution $p(x|y)$, denoted as $\dkl{p(x|y)}{p(x)}$, is an effective tool for both dataset-level and image-level distribution shift detection.
Intuitively, the more the measurement $y$ ``pulls" the posterior away from the prior, the higher the chance that $x$ is OOD. As a more concrete motivation, consider a toy example of a Gaussian prior $p(x)=\mathcal{N}(0,\sigma_1^2)$, with a forward model $y=x+\epsilon$ under the noise model $\epsilon\sim\mathcal{N}(0, \sigma_2^2)$. The posterior in this example is also a Gaussian, with mean proportional to the measurement $y$:
\begin{equation}
\label{eq:toy_posterior}
    p(x|y)=\mathcal{N}\big(\frac{\sigma_1^2}{\sigma_1^2+\sigma_2^2}y,\frac{\sigma_1^2\sigma_2^2}{\sigma_1^2+\sigma_2^2}\big).
\end{equation}
The KL-divergence between the prior and the posterior is
\begin{equation}
\label{eq:toy_kl_divergence}
    D_{KL}\bigl(p(x|y)||p(x)\bigr)=\frac{\sigma_1^2y^2}{2(\sigma_1^2+\sigma_2^2)^2} + \text{constant}.
\end{equation}
This KL divergence grows quadratically with $|y|$.
Under the forward model in this example, a signal $x^\star$ that gives rise to an observation $y$ with a larger magnitude is increasingly unlikely under the prior, so the KL-divergence effectively captures where $x^\star$ lies relative to the prior distribution.

In general, the KL divergence between the prior and posterior distributions is not always easily accessible.
However, under certain regularity conditions \cite{song2021maximum}, the KL divergence of two arbitrary distributions $\dkl{p}{q}$ can be written in terms of the scores of the marginal distributions $p_t$ and $q_t$, under a fixed SDE of the form in \Cref{eq:forward_sde}.
Specifically, 
\begin{equation}
\label{eq:dkl}
\dkl{p}{q} =\frac{1}{2}\int_{0}^TE_{x \sim p_t} \Bigl[ \|g(t)h(x,t)\|_2^2 \Bigr]dt ,
\end{equation}
where $p_0 = p$, $q_0 = q$, $g(t)$ is the drift of the SDE, and
\begin{equation}
\label{eq:h_t}
h(x,t)=\nabla_x\log{p_t(x)} - \nabla_x\log{q_t(x)} .
\end{equation}
While \Cref{eq:dkl} also cannot be computed in general unless the two scores in \Cref{eq:h_t} are available, we observe that posterior sampling offers access to estimates of these scores for the prior and the posterior.
Replacing $p$ and $q$ with the posterior distribution $p(x|y)$ and the unconditional training data distribution $p(x)$, respectively, following \Cref{eq:posterior_score}, $h(x,t)$ becomes the score of the likelihood $\nabla_x\log{p_t(y|x)}$.
As described in \Cref{sec:preliminaries}, posterior sampling estimates this score with $s_l(x,y;t)$, which yields the following approximation of the KL divergence:
\begin{equation}
    \label{eq:dkl_posterior}
    \begin{split}
    D_{KL}\bigl(p(x|y&)||p(x)\bigr)=\\
    &\frac{1}{2}\int_{0}^TE_{x \sim p_t(x|y)} \Bigl[ \|g(t)s_l(x,y;t)\|_2^2 \Bigr]dt .
    \end{split}
\end{equation}
To compute this KL divergence given measurement $y$, we first generate multiple samples following the discrete posterior sampling process described in \Cref{eq:posterior_sampling}, each with 
different
random $z\in \mathcal{N}(0,I)$.
Along the way, we collect the estimated likelihood scores $s_l(x_t,y;t)$ corresponding to each noise $z$ and each timestep $t$.
Afterwards, we average the norm in \Cref{eq:dkl_posterior} over random samples to approximate the expectation, and sum over timesteps, to approximate the integral in \Cref{eq:dkl_posterior}.

\paragraph{\name{}: \Block{} \& Timestep Restriction.}
To effectively detect and identify spatially localized OOD image features, we consider the restriction of the Kullback-Leibler (KL) divergence in both space and time.

First, we divide each score $s_l(x,y;t)\in\mathbb{R}^{D\times D}$ into a total of $N_B$ \blocks{}, each with size $D_B\times D_B$. We denote the restriction of the score to each \block{} $B_i$ as $s_l(x,y;t)\big|_{B_i}$.
This is motivated by a longstanding statistics and information-theory literature on non-parameteric estimation of the KL divergence using histogram or data-dependent partitions, which represents the divergence as an aggregation of cell-wise contributions \cite{wang2005divergence, silva2010information, paninski2003estimation}.
Since \Cref{eq:dkl_posterior} uses a squared $l_2$ norm over spatial coordinates, restricting to blocks yields a localized contribution.
This helps mitigate signal dilution when the OOD feature occupies a small region, by preventing the local distribution shift from being averaged out by the rest of the image, as shown in \Cref{fig:method_motivation}.
Analogous to the histogram bin widths in previous works, block size $D_B$ controls the trade-off between better localization and higher variance.

Second, we further restrict the timestep ranges to integrate over in \Cref{eq:dkl_posterior}.
For each \block{} $B_i$, we define $\text{KLIP}(B_i,[t_0,t_1];y)$ by restricting \Cref{eq:dkl_posterior} to that spatial block as well as the timestep range $[t_0, t_1]$:
\vspace{-2mm}
\begin{align}
\text{KLIP}&(B_i,[t_0,t_1];y)=\nonumber \\
&\frac{1}{2}\int_{t_0}^{t_1}E_{x \sim p_t(x|y)} \Bigl[ \|g(t)s_l(x,y;t)\big|_{B_i}\|_2^2 \Bigr]dt , \label{eq:dkl_block}
\end{align}
where $t_0<t_1$.
This time-range restriction is inspired by previous works \cite{chen2024exploring, rissanen2023generative, Spectraldiffusion} studying the different roles each timestep plays during sampling, including generating features at different scales \cite{rissanen2023generative, Spectraldiffusion} and modeling images over a lower-dimensional manifold at certain timesteps \cite{chen2024exploring}.

The benefits of these restrictions are demonstrated in \Cref{fig:method_motivation}. 
The distributions of whole image prior--posterior KL divergence for ID images (normal CT scans) and OOD images (CT scans with small synthetic star artifacts), shown in \Cref{fig:method_motivation} (b), do not display any separation.
In contrast, the distributions of \block{}-restricted KL divergence for \blocks{} containing and not containing the OOD star artifact, shown in \Cref{fig:method_motivation} (c), are much better separated.
Similarly, timestep restriction can reveal patterns that are informative for distribution shift detection.
\Cref{fig:method_motivation} (d) shows that when integrated over a sliding timestep window of size $0.05$ (relative to $T=1$), OOD blocks exhibit a bump at around $t=0.3$, that is absent in ID blocks (see solid lines).
This pattern is not captured when integrating over the entire timestep range $[0,1]$ (see dotted lines).

\begin{table*}[ht]
    \centering
    \caption{\textbf{AUC results for dataset-level and image-level OOD detection}, for different OOD artifacts, models, and inverse problems. We mark with $\dagger$ those AUC values that correspond to settings for which we tuned the \name{} hyperparameters. Other experiments use the same set of hyperparameters without further refinement. The best AUC for each task and dataset is underlined. Note that CutPaste and SimpleNet operate directly on images rather than measurements, making their OOD detection task somewhat easier.}
    \vspace{-2mm}
    \label{tab:main_results_song}
    \footnotesize
    \setlength{\tabcolsep}{3pt}
    \begin{tabular}{lcccccccccccclcc}
        \toprule
            & \multicolumn{10}{c}{\textbf{Predictor-Corrector \cite{song2021solving}}}
            & \multicolumn{2}{c}{\textbf{PaDIS \cite{padis}}}
            &
            & \multicolumn{2}{c}{\textbf{DDPM \cite{diffusers}}}
            \\
        \cmidrule(lr){4-9}  \cmidrule(lr){12-13} \cmidrule(lr){15-16}

            & \multicolumn{6}{c}{\textbf{Dataset (CT)}}
            & \multicolumn{6}{c}{\textbf{Image (CT)}}
            &
            & \multicolumn{2}{c}{\textbf{Image (Deblur)}}
            \\
        \cmidrule(lr){2-7} \cmidrule(lr){8-13} \cmidrule(lr){15-16}
            & CutPaste
            & SimpleNet 
            & NLL
            & DiffPath
            & $D_{KL}$
            & \name{}
            & $D_{KL}$
            & \name{}
            & CutPaste 
            & SimpleNet
            & $D_{KL}$
            & \name{}
            &
            & $D_{KL}$
            & \name{}
            \\
        \cmidrule(lr){1-13} \cmidrule(lr){14-16}
        \textbf{Star}
            & \underline{\songAlgOneStar}
            & \songAlgTwoStar
            & \songNLLstar                  
            & \songDiffpathstar
            & \songDKLstar
            & {\songoursstar}$^\dagger$
            & \songDKLstarimg
            & {\songoursstarimg}$^\dagger$
            & \songAlgOneStarimg
            & \underline{\songAlgTwoStarimg}
            & \PaDISDKLstarimg
            & {\PaDISoursstarimg}
        &\textbf{Scar}
            & \CelebADKLsyntheticimg
            & \underline{\CelebAourssyntheticimg}
            \\
        \textbf{Tumor}
            & \songAlgOneTumor
            & \songAlgTwoTumor
            & \songNLLtumor                  
            & \songDiffpathtumor
            & \songDKLtumor
            & \underline{\songourstumor}
            & \songDKLtumorimg
            & \underline{\songourstumorimg}
            & \songAlgOneTumorimg
            & \songAlgTwoTumorimg
            & \PaDISDKLtumorimg
            & {\PaDISourstumorimg}
        &\textbf{Character}
            & \CelebADKLcollectedimg
            & \underline{\CelebAourscollectedimg}
            \\
        \bottomrule
    \end{tabular}
    \vspace{-4mm}
\end{table*}

\begin{figure*}[t]
  \centering

    \includegraphics[width=1.0\textwidth]{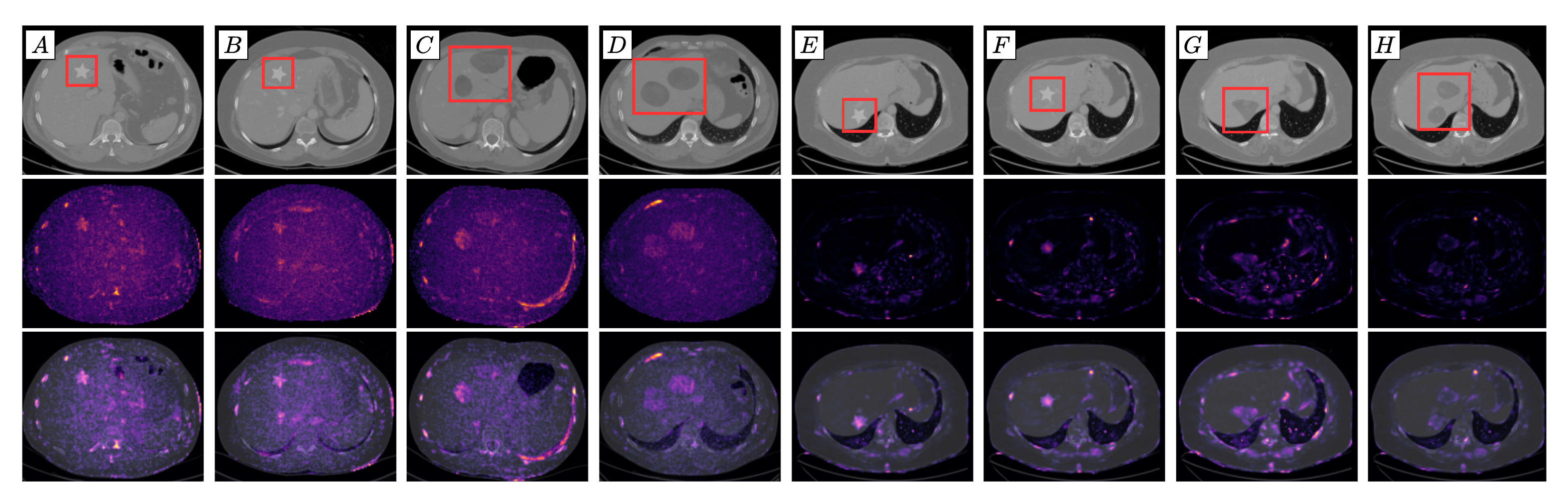}
    \vspace{-5mm}
    \caption{\textbf{Visualization of image-level OOD detection in sparse-view CT scans.} \emph{Top:} Images in the \calibrationset{} and \oodset{}. Red boxes annotate the stars and tumors. \emph{Middle:} Heatmaps of \name{} computed pixel-wise (i.e. $D_B=1$). \emph{Bottom:} Heatmaps overlaid on images, to show localization. 
    Columns A through D use a predictor-corrector based diffusion model, and columns E through H use a patch-based diffusion model. For each model, we show two examples from the \calibrationset{} (stars) and two from the \oodset{} (tumors).
    }
    \vspace{-4mm}
   \label{fig:combined_image_level}
\end{figure*}

\begin{figure*}[t]
  \centering

    \includegraphics[width=1.0\textwidth]{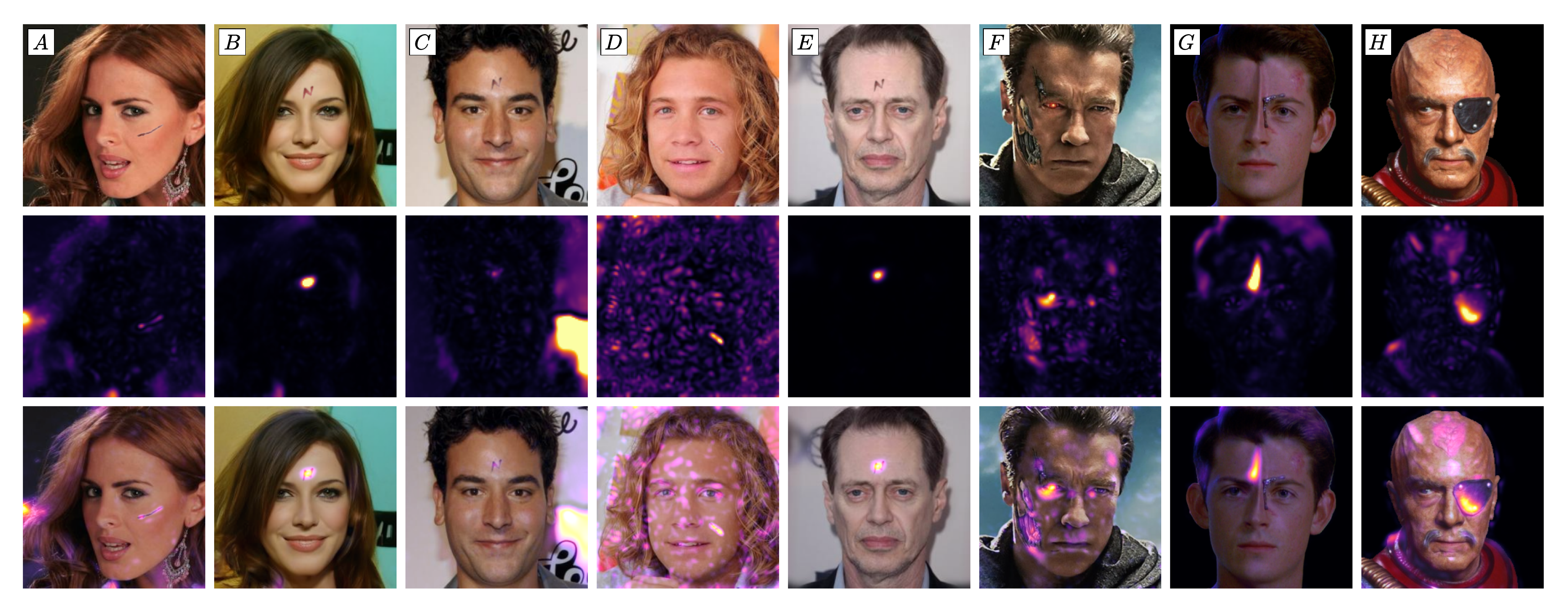}
    \vspace{-6mm}
    \caption{\textbf{Visual results for image-level OOD detection on human faces.} \emph{Top:} Samples of the \oodset{} containing different localized artifacts such as scars. \emph{Middle:} Heatmaps of \name{} computed pixel-wise (i.e. $D_B=1$). \emph{Bottom:} Heatmaps overlaid on images, to show localization. Columns A through E are images from the CelebA test set with synthetically-added scars; columns F through H are images from film and television of characters with distinctive and localized facial features.}
    \vspace{-2mm}
   \label{fig:celebA}
\end{figure*}

\section{Evaluation}
\label{sec:evaluation}

\paragraph{Models \& Datasets.}
We evaluate the effectiveness of our proposed OOD metric KLIP (\Cref{eq:dkl_block}), across different settings.
We first consider computed tomography (CT) as our inverse problem, for which we consider two distinct diffusion models: a predictor-corrector based model \cite{song2020score, song2021solving} and a patch-based model (PaDIS) \cite{padis}.
We selected these models to assess the generalizability of \name{}, as these models
differ not only in the sampling algorithm, but also in how the likelihood score is approximated.
Further details on the approximations involved in each method for posterior sampling can be found in \Cref{sec:sup_likelihood}.

We define the inlier distribution as healthy abdominal CT scans, and liver tumors 
as the local OOD features absent in the inlier distribution.
We train both diffusion models on the training set of the Combined Healthy Abdominal Organ Segmentation (CHAOS) dataset \cite{CHAOS}, which contains 512 $\times$ 512 resolution abdominal CT scans of 40 different patients.
Since the patients whose CT scans are included in the dataset are potential liver donors with healthy livers, liver tumors are guaranteed to be both localized and out of distribution.
We select a subset of 100 images from the test set as the \idset{}, and generate an \oodset{} by adding realistic synthetic tumors to each image of the \idset{}.
We generate synthetic liver tumors following \cite{synthetic_tumor}, whose results closely resemble the tumors found in real patients.

We also validate the localization of OOD features on another inverse problem, Gaussian de-blurring of face images.
For this task, we use CelebA \cite{CelebA} as the inlier distribution, and perform Gaussian de-blurring with a DDPM-based \cite{DDPM} diffusion model \cite{diffusers} trained on that dataset.
We generate two locally OOD datasets: one by adding realistic synthetic scars of various shapes to the faces of 20 different CelebA test images, and another by collecting 15 images of human-like movie and television characters, including those from Star Trek and Harry Potter, whose faces contain localized unusual features not present in the ID CelebA dataset \cite{CelebA}.

\vspace{-3mm}
\paragraph{Experiments.}
We first evaluate the performance of \name{} on distinguishing OOD and ID CT scans, based on their projection measurements.
For each image in the \idset{} and the \oodset{}, we simulate sparse-view CT measurements (with 24 projection angles) and compute \name{}, approximating the expectation in \Cref{eq:dkl_block} with the average over 5 samples for each measurement. 
For image-level OOD detection we use only the images in the \oodset{}, and mask out the parts of CT scan outside the body mass.
We repeat the image-level experiment with CelebA \cite{CelebA} as inlier distribution by using \name{} to localize the OOD features on human and human-like faces.
On this dataset, the expectation is estimated over 8 samples instead of 5.
In \Cref{sec:sampling} we report an ablation study over the number of samples used to approximate the expectation.

We note that \name{} has multiple hyperparameters: the \block{} size $D_B$ and the time ranges $[t_0,t_1]$.
To choose these values, we generate a \calibrationset{} by adding synthetic star-shaped artifacts to each CT scan image in the \idset{}.
Then, we perform a grid search over the hyperparameters to maximize OOD detection performance with respect to this \calibrationset{} using the predictor-corrector based model \cite{song2021solving}. 
We tune hyperparameters separately for each task (dataset-level and image-level OOD detection) but use the same hyperparameter values for experiments on the more realistic tasks of tumor detection in sparse-view CT and face artifact detection in Gaussian de-blurring, for all models.

\vspace{-3mm}
\paragraph{Baselines \& Metrics.}

We compare \name{} against existing OOD detection techniques, across both dataset-level and image-level tasks.
For diffusion model based techniques, we compare against DiffPath (6D) \cite{diffpath} and the Negative Log likelihood (NLL), via its Evidence Lower Bound (ELBO) when NLL cannot be acquired; these methods are only suitable for dataset-level OOD detection.
We closely follow the official implementations of \cite{diffpath} and \cite{song2020score} to compute DiffPath (6D) and NLL respectively; full details of these implementations are provided in \Cref{sec:sup_baseline} and in our code.

We also compare against the raw KL divergence in \Cref{eq:dkl_posterior} (denoted as $D_{KL}$), without our timestep restrictions; $D_{KL}$ treats the entire image as one block for dataset-level OOD detection and considers each pixel as a separate block for image-level OOD detection.
Additionally, we compare against two state of the art methods for anomaly detection that are not based on diffusion models: CutPaste \cite{cutpaste} and SimpleNet \cite{simplenet}. We emphasize that CutPaste and SimpleNet operate directly on the true image, rather than taking indirect (CT or blurred) measurements as the diffusion-based methods do; this makes their OOD detection task somewhat easier.

We report OOD detection performance using the Area Under the receiver-operating characteristic Curve (AUC). 
For the dataset-level OOD detection task, we consider an image to be OOD if its \name{} value exceeds a threshold at any \block{} of the image. 
For the image-level OOD detection task, we consider a \block{} to be OOD if its \name{} value exceeds a threshold
and report AUC evaluated over all \blocks{} across all images in the \oodset{}.

\vspace{-3mm}
\paragraph{Results.}

Our main results are reported in \Cref{tab:main_results_song}; computation time is reported in \Cref{sec:computationalcost}.
\name{} applied to the predictor-corrector based diffusion model \cite{song2021solving} outperforms all baselines for both dataset and image-level OOD detection, across both the star artifacts in the \calibrationset{} and the realistic synthetic liver tumors in the \oodset{}, in the context of sparse-view CT reconstruction.
This localization capability extends to the patch-based diffusion model \cite{padis} using the same hyperparameters, and to detection of both scars and character features in the context of human face de-blurring with a DDPM-based model \cite{diffusers}.

\Cref{fig:combined_image_level} visualizes examples of OOD feature localization using \name{}, for CT scans in the \calibrationset{} (with stars) and \oodset{} (with tumors) and for both the predictor-corrector based model (first 4 columns) and the patch-based model (last 4 columns).
\name{} heatmaps largely coincide with the synthetic stars and tumors in the CT scans.
While some \blocks{} that do not contain OOD features also have high \name{} values, these values are consistently high for \blocks{} that do contain local OOD features.

Similarly, \Cref{fig:celebA} visualizes examples of OOD feature localization using \name{} with a DDPM-based model \cite{diffusers} trained on CelebA \cite{CelebA}, for Gaussian de-blurring of human faces with localized unusual (OOD) features. The first five columns show images from the CelebA test set, with added scars (inspired by Harry Potter's); for these examples we can ensure that only the added scar is OOD. The rightmost three columns show examples of film and television characters with unusual facial features such as facial prosthetics. No postprocessing is done on these images, but they may contain subtle image-wide distribution shifts compared to CelebA, in addition to the localized facial prosthetics.
Across both types of face OOD detection, we find that \name{} successfully localizes the OOD features, though in some cases it also highlights other parts of the image.
\vspace{-5mm}

\begin{figure*}[t]
  \centering
    \includegraphics[width=1.0\textwidth]{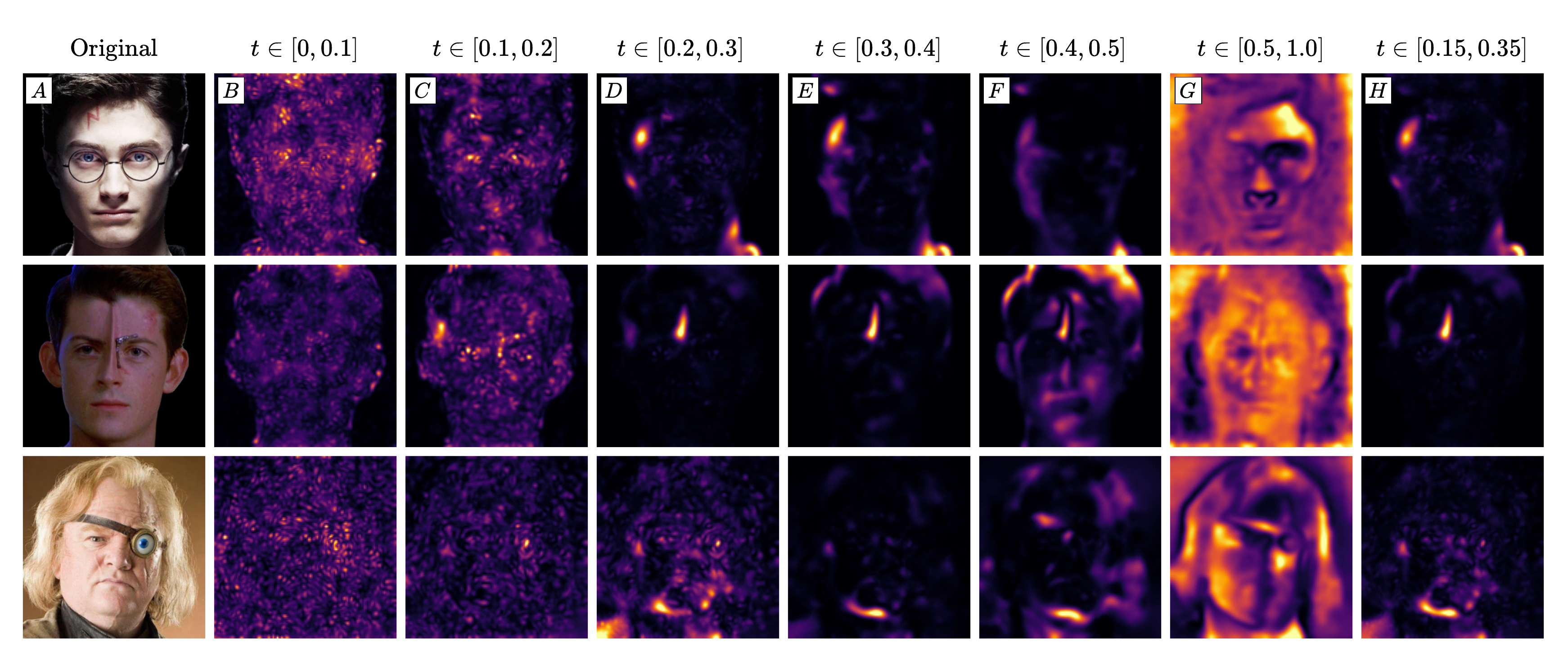}
    \vspace{-5mm}
    \caption{\textbf{Effect of different timestep ranges on KLIP.} 
    (A) Original images before Gaussian blur is applied. (B-F) KLIP heatmaps for timestep windows of length 0.1, $[t_0, t_0+0.1]$, where $t_0$ starts at 0 in (B) and is incrementally increased in steps of 0.1 up to 0.4.
    (G) KLIP heatmap for the window $t \in [0.5, 1.0]$.
    (H) KLIP heatmap for the window $t \in [0.15, 0.35]$, which was tuned to optimize detection of star artifacts in sparse-view CT using a different diffusion model. We observe that different features are detected at different timesteps, in line with expectations that diffusion models add higher-frequency details later in sampling (at smaller values of $t$) \cite{rissanen2023generative, Spectraldiffusion}.}
    \vspace{-3mm}
   \label{fig:celebA_ablation}
\end{figure*}

\paragraph{Ablation Study.}
We perform three ablation experiments. 
First, we show that both \block{} and timestep restrictions of \Cref{eq:dkl_posterior} are necessary to achieve both \emph{dataset-level} and \emph{image-level} detection of local OOD features.
As in our main experiments, we select hyperparameters that maximize performance over the \calibrationset{} (with stars) and use the same values on the \oodset{} (with tumors).
\Cref{tab:ablation_restriction} summarizes the resulting AUCs.
We observe that restrictions to \blocks{} or timestep ranges individually tend to improve performance, and that both restrictions (i.e., \name{}) usually perform better than either restriction alone.

\begin{table}[t]
    \centering
    \caption{\textbf{Ablation study over separate \block{} and timestep restriction.} All AUC values are reported for the predictor-corrector based model. Hyperparameters are optimized over the \calibrationset{} (star); the corresponding tuned AUC values are marked with $\dagger$. The best AUC for each task is underlined.} 
    \vspace{-2mm}
    \label{tab:ablation_restriction}
    \begin{tabular}{lcccc}
        \toprule
                & \multicolumn{2}{c}{\textbf{\emph{Dataset-Level}}}
                & \multicolumn{2}{c}{\textbf{\emph{Image-Level}}}
                \\
        \cmidrule(lr){2-3} \cmidrule(lr){4-5}
                & Star
                & Tumor
                & Star
                & Tumor
                \\
        \midrule 
            $D_{KL}$
                & 0.54$^\dagger$
                & 0.60
                & 0.84$^\dagger$
                & 0.86
            \\
            \hspace{1mm} + \block{}
                & 0.85$^\dagger$
                & 0.65
                & 0.88$^\dagger$
                & \underline{0.91}
            \\
            \hspace{1mm} + time
                & 0.57$^\dagger$
                & \underline{0.78}
                & 0.86$^\dagger$
                & 0.81
            \\
            \name{}
                & \underline{0.86$^\dagger$}
                & \underline{0.78}
                & \underline{0.91$^\dagger$}
                & 0.88
            \\
        \bottomrule
        \vspace{-5mm}
    \end{tabular}
\end{table}

\begin{table}[t]
    \centering
    \caption{\textbf{Sensitivity to hyperparameters.} Our main results are obtained by tuning hyperparameters on the \calibrationset{} (stars) for the predictor-corrector based diffusion model. Here we compare AUC values when hyperparameters are tuned instead on the \oodset{} (tumors), to evaluate hyperparameter sensitivity. Tuned values are marked with $\dagger$. The best AUC for each task is underlined.}
    \vspace{-2mm}
    \label{tab:ablation_sensitivity}
    \begin{tabular}{lcccc}
        \toprule
                & \multicolumn{2}{c}{\textbf{\emph{Dataset-Level}}}
                & \multicolumn{2}{c}{\textbf{\emph{Image-Level}}}
                \\
        \cmidrule(lr){2-3} \cmidrule(lr){4-5}
                & Star 
                & Tumor
                & Star
                & Tumor
                \\
        \midrule 
            Tune on star
                & \underline{0.86$^\dagger$}
                & 0.78
                & \underline{0.91$^\dagger$}
                & 0.88
            \\
            Tune on tumor
                & 0.68
                & \underline{0.87$^\dagger$}
                & 0.85
                & \underline{0.92$^\dagger$}
            \\

        \bottomrule
        \vspace{-10mm}
    \end{tabular}
\end{table}

Second, we present a sensitivity analysis of AUC performance with respect to the choice of \block{} size and timestep range.
\Cref{tab:ablation_sensitivity} shows how the AUCs change when the hyperparameters are optimized to maximize performance over the \oodset{} instead of the \calibrationset{}.
We find that tuning performance on one OOD dataset can reduce performance on the other; future investigation into robust hyperparameter selection should improve performance even further.

Third, in
\Cref{fig:celebA_ablation}, we qualitatively visualize \name{} for three OOD face images across varying timestep ranges
and observe that different time ranges capture OOD features of varying sizes and shapes.
At the earliest timesteps (large $t$s), \name{} highlights low-frequency components of the image such as the overall shape of the face.
Smaller, more localized OOD features become more prominent as sampling proceeds (small $t$s).
For example, Harry Potter's scar (top row), Icheb's metal implant (middle row), and Mad-Eye Moody's eye (bottom row) are all visible in panel C, while Moody's eye and Icheb's ridged nose, which are both larger, first appear earlier in panels D and F, respectively.
This aligns with the observations of previous works \cite{Spectraldiffusion, rissanen2023generative} that diffusion models construct low-frequency  signals in the early stage of sampling, and progressively add fine details throughout sampling.
It also explains why the hyperparameters $[t_0,t_1]$ that best detect local distribution shifts depend on the characteristics and scale of the OOD features.

\section{Discussion}
\label{sec:conclusion}

We focus on the understudied and challenging task of detecting spatially localized OOD features in images, given access only to indirect measurements of those images in the context of an inverse problem. 
We propose an OOD detection metric, \name{}, based on the spatially and temporally restricted KL divergence between the prior and posterior distributions, which we can estimate efficiently during posterior sampling with a diffusion model. 
This metric outperforms existing diffusion-based OOD detection methods across diverse imaging inverse problems (sparse-view CT and Gaussian deblurring) and inlier distributions (abdominal CT scans and human faces), for both dataset-level and image-level OOD detection of spatially localized OOD features. It also transfers across diffusion models, including both whole-image and patch-based architectures.

\vspace{-3mm}
\paragraph{Limitations and Future Work.}
While our results are promising, they are also subject to the following limitations.
First, \name{} is sensitive to hyperparameters for block size and time window, and could likely be improved with a more refined tuning strategy. We posit that this sensitivity is due to variation in the size and shape of real localized OOD artifacts, which influences the optimal block sizes and time windows for detection.
Second, our evaluation framework is technically an \emph{inverse crime}, because we use the same forward models to simulate the measurements (CT projections and blurred images) and perform posterior sampling reconstruction. While this provides a fair testbed to compare different OOD detection methods, and we show some robustness to model misspecification in \Cref{sec:modelmismatch}, we hope that future work will validate and refine \name{} on real-world data for diverse local OOD detection tasks.

\clearpage
\section*{Acknowledgment}
This work was supported in part by the HIVES program at Georgia Tech Research Institute (GTRI).

{
    \small
    \bibliographystyle{ieeenat_fullname}
    \bibliography{main}
}
\clearpage
\setcounter{page}{1}
\maketitlesupplementary

\section{Likelihood Score Approximation}
\label{sec:sup_likelihood}
We evaluate KLIP in the context of two prior works \cite{song2021solving, dps} with different posterior sampling algorithms. In \cite{song2021solving}, an additional step at each time $t$ replaces the sample $x_t$ with $x_t'$, which is the solution of a proximal optimization step to ensure consistency of the sample with the measurement $y$. Specifically, $x_t'$ is the solution of the optimization problem
\begin{equation}
\begin{split}
x_t'=\arg\min_{z\in\mathbb{R}^D}\{(1-\lambda)\|z-x_{t}\|_B^2+\min_{u\in\mathbb{R}^D}&\|z-u\|_B^2\}\\
&s.t.\hspace{2mm}Au=y_t,
\end{split}
\end{equation}
where $A$ is a linear forward model that involves an invertible square matrix $B$, $y_t$ is the simulated measurement using the current sample $x_t$, and $\lambda$ is a hyperparameter that controls how strongly the measurements should affect the sampling process. We refer interested readers to \cite{song2021solving} for more details. While this work does not directly approximate the likelihood score, the update made by the proximal step is precisely the difference between unconditional sampling (\Cref{eq:backward_sampling}) and posterior sampling (\Cref{eq:posterior_sampling}), which can be expressed as the following equation:
\begin{equation}
-g(t)^2s_l(x_t,y;t)\simeq x_t'-x_t.
\end{equation}
Therefore, we consider the scaled update
\begin{equation}
s_l(x_t,y;t)\simeq\frac{x_t-x_t'}{g(t)^2}
\end{equation}
as an implicit approximation of the likelihood score, and use it to compute \name{}.

On the other hand, \cite{dps} approximates the score with
\begin{equation}
\label{eq:dps_likelihood}
\nabla_{x_t}\log{p(y|x_t)}\simeq-\frac{1}{\sigma^2}\nabla_{x_t}\| y - \mathcal{A}(\hat{x}_0(x_t)) \|_2^2,
\end{equation}
where $\mathcal{A}$ is the forward model, $\hat{x}_0(x_t)=\mathbb{E}[x_0|x_t]$ is the predicted state at $t=0$ given the current state $x_t$, $\sigma^2$ is the variance of the Gaussian noise included in the forward model, and $\zeta_t$ is a parameter that controls how strong the likelihood affects the sampling process.

\section{Baseline Computation}
\label{sec:sup_baseline}

We compare \name{} against 2 primary baselines, NLL (negative log likelihood) and DiffPath \cite{diffpath}. For NLL, we directly use the official implementation on github by the authors of \cite{song2020score}. It computes the exact likelihood instead of the Evidence Lower Bound (ELBO) using the probability flow ODE, which is an ordinary differential equation whose marginals coincide with the underlying SDE of the diffusion model. 

DiffPath \cite{diffpath} constructs a 6-dimensional feature vector for each image, and uses these features to determine if the image is OOD.
These features are the $\ell_1$, $\ell_2$, and $\ell_3$
norms of $\epsilon_\theta(x_t,t)$, defined as:
\begin{equation}
\label{eq:diffpath_eps}
\epsilon_\theta(x_t,t)=-\sigma(t)s_\theta(x_t,t) ,
\end{equation}
and its time derivative.
Here, $\sigma(t)$ is a function used to define the diffusion coefficients $g(t)$ of the SDE.
Specifically, both diffusion models that we evaluate \name{} on are defined by a class of SDEs called Variance Exploding (VE) SDEs, which take the form
\begin{equation}
\label{eq:ve_sde}
dx=\sqrt\frac{d\sigma^2(t)}{dt}dw.
\end{equation}
To evaluate $\epsilon_\theta(x_t,t)$, the score model is queried once every timestep.
Since the predictor-corrector based model \cite{song2021solving} involves multiple updates by the predictor and the corrector at each timestep, we query the score model once after the last corrector update for each timestep.
Additionally, DiffPath fits a Kernel Density Estimator (KDE) to the feature vectors of the images in the training set. 
For both models, we instead fit DiffPath's KDE to feature vectors extracted from images in the ID validation set, that do not overlap with images in the ID training or evaluation sets.
We used 250 images to fit KDE for the predictor-corrector model, and 100 images for the patch-based model.

\section{Robustness Evaluation}
\label{sec:sup_ablation}
\begin{figure*}[ht]
  \centering
    \includegraphics[width=1.0\textwidth]{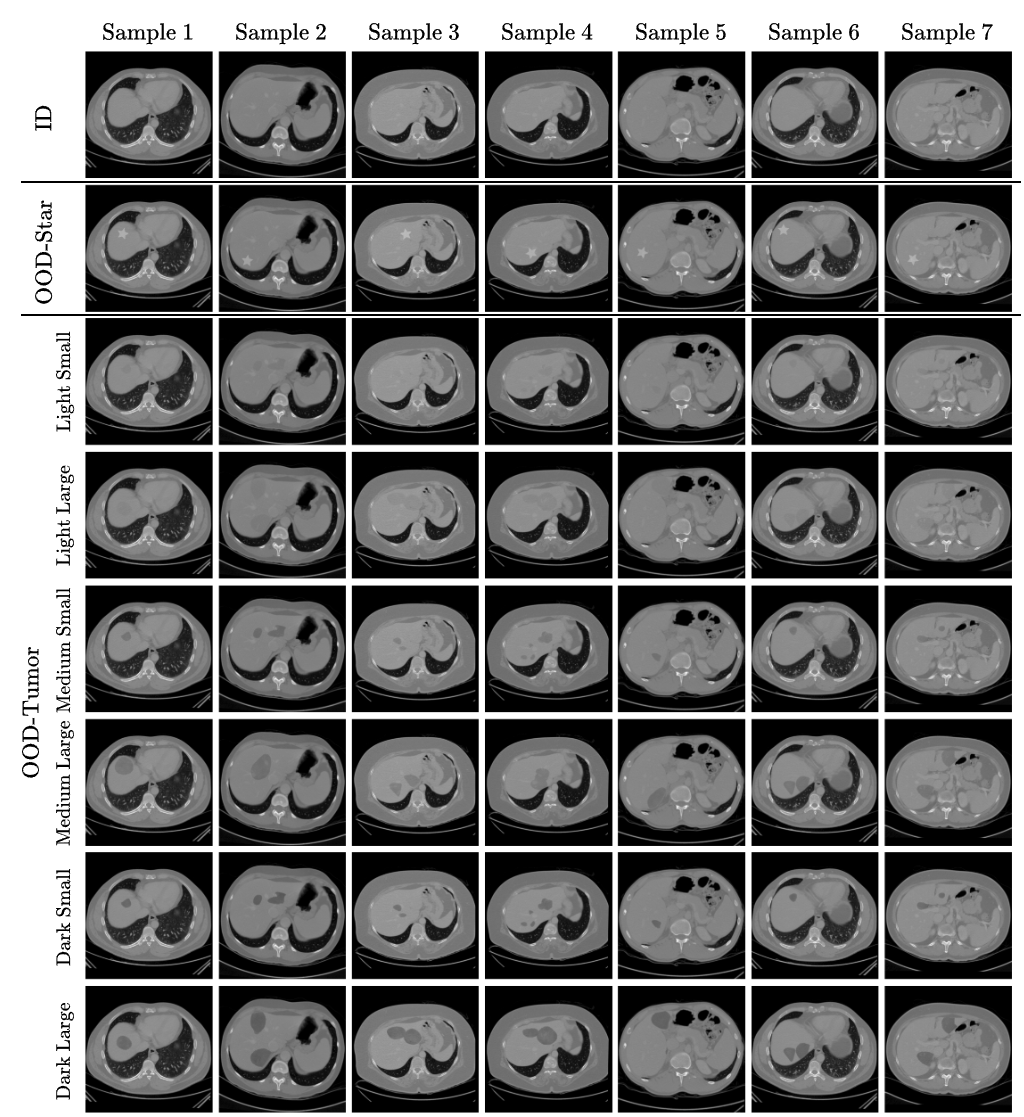}
    \vspace{-5mm}
    \caption{\textbf{Sample images included in the ID and different OOD sets.} Each row represents a distinct dataset, and each column shows 7 different samples from that dataset. From the top, we have (1) the ID set consisted of healthy CT scans from the CHAOS \cite{CHAOS} evaluation dataset, (2) the OOD set with a star shaped artifact, used for hyperparameter tuning, and (3) 6 OOD sets containing synthetic liver tumors of different darkness and shape, generated following \cite{synthetic_tumor}.}
    \vspace{-2mm}
   \label{fig:sup_samples}
\end{figure*}

We perform an additional comparison to evaluate the generalizability of \name{} to local OOD features with different properties, specifically simulated liver tumors with different sizes and densities.
\Cref{fig:sup_samples} shows sample images from the different OOD sets we used for this evaluation, acquired by modifying the darkness and size of the simulated liver tumors.
The OOD set that we used for evaluation in the main manuscript corresponds to the 6th row, containing large tumors with medium darkness.

We also compare \name{} against two additional baselines: CutPaste \cite{cutpaste} and SimpleNet \cite{simplenet}, using their public code and default parameter settings (with the exception of image size, which we adjust to match our dataset).
While these works mainly aim to detect anomalies or defects in the context of industrial visual inspection, they have been extended to other settings including medical imaging \cite{anomaly1, anomaly2, anomaly3}. We train both CutPaste and SimpleNet with the training set of the CHAOS dataset \cite{CHAOS}, which we used to train both of our diffusion models.

The results are presented in \Cref{tab:sup_ablation}. 
We find that dataset-level OOD detection performance is fairly stable across tumor types, for all models. \name{} yields the highest dataset-level AUC across all tumor types, when using the predictor-corrector based diffusion model. However, all dataset-level OOD metrics struggle when applied to the patch-based PaDIS diffusion model. Empirically, we observe that \name{} can achieve higher AUC metrics for PaDIS when its hyperparameters are chosen specifically for that model, so we are optimistic that future work may introduce a more adaptive hyperparameter selection strategy that would render \name{} more robust across different diffusion model architectures.

\begin{figure*}[ht]
  \centering
    \includegraphics[width=1.0\textwidth]{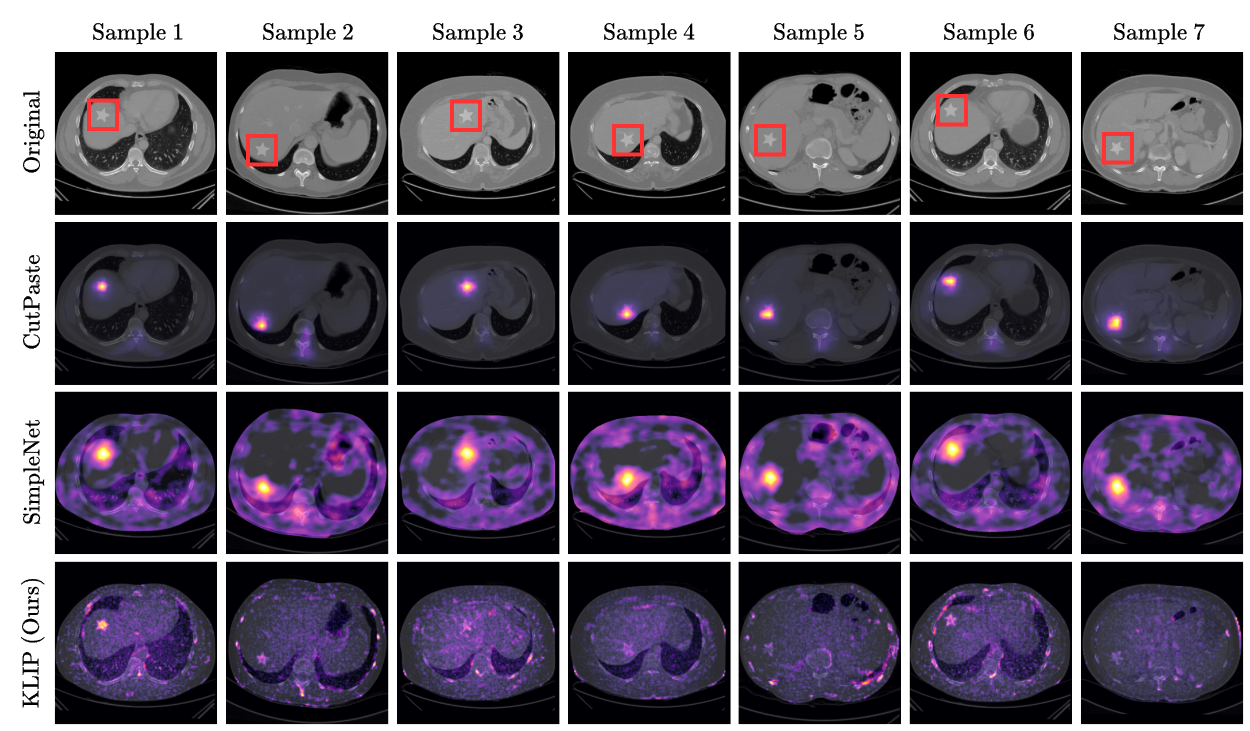}
    \vspace{-5mm}
    \caption{\textbf{Visual results for image-level OOD detection on sparse-view CT scans.} 
    \textit{Row 1:} Images in the OOD set with synthetic star artifacts. Red boxes annotate where the stars are.
    \textit{Rows 2-4:} Heatmaps of CutPaste \cite{cutpaste}, SimpleNet \cite{simplenet}, and KLIP overlaid on images.}
    \vspace{-4mm}
   \label{fig:sup_star}
\end{figure*}

\begin{figure*}[ht]
  \centering
    \includegraphics[width=1.0\textwidth]{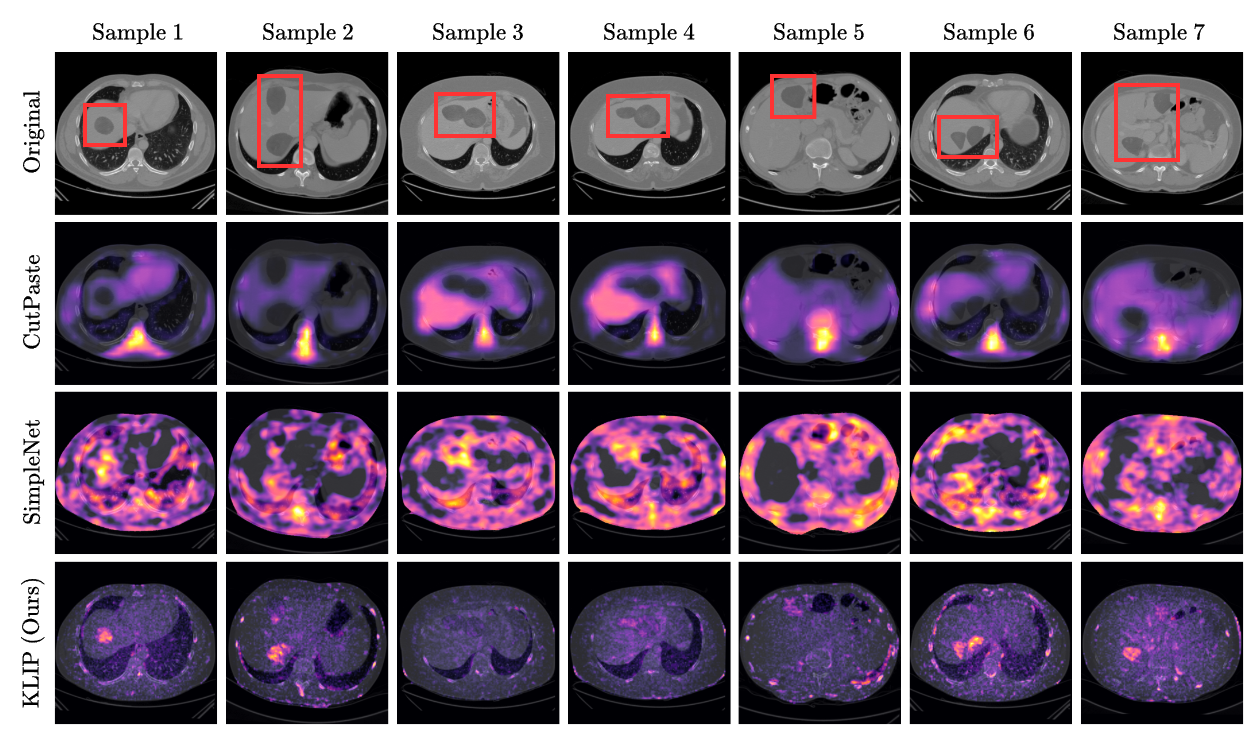}
    \vspace{-5mm}
    \caption{\textbf{Visual results for image-level OOD detection on sparse-view CT scans.} 
    \textit{Row 1:} Images in the OOD set with dark and large tumors. Red boxes annotate where the tumors are.
    \textit{Rows 2-4:} Heatmaps of CutPaste \cite{cutpaste}, SimpleNet \cite{simplenet}, and KLIP overlaid on images.}
    \vspace{-4mm}
   \label{fig:sup_tumor_dark_large}
\end{figure*}

\begin{figure*}[ht]
  \centering
    \includegraphics[width=1.0\textwidth]{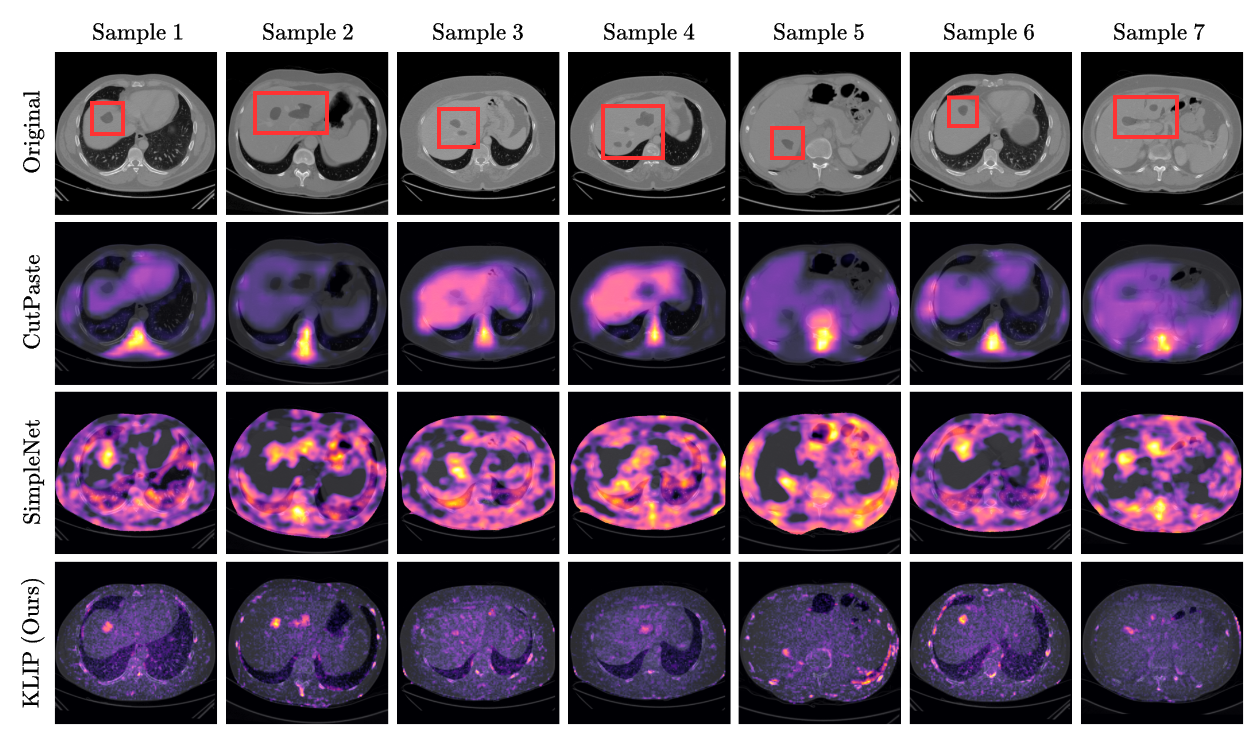}
    \vspace{-5mm}
    \caption{\textbf{Visual results for image-level OOD detection on sparse-view CT scans.} 
    \textit{Row 1:} Images in the OOD set with dark and small tumors. Red boxes annotate where the tumors are.
    \textit{Rows 2-4:} Heatmaps of CutPaste \cite{cutpaste}, SimpleNet \cite{simplenet}, and KLIP overlaid on images.}
    \vspace{-4mm}
   \label{fig:sup_tumor_dark_small}
\end{figure*}

\begin{figure*}[ht]
  \centering
    \includegraphics[width=1.0\textwidth]{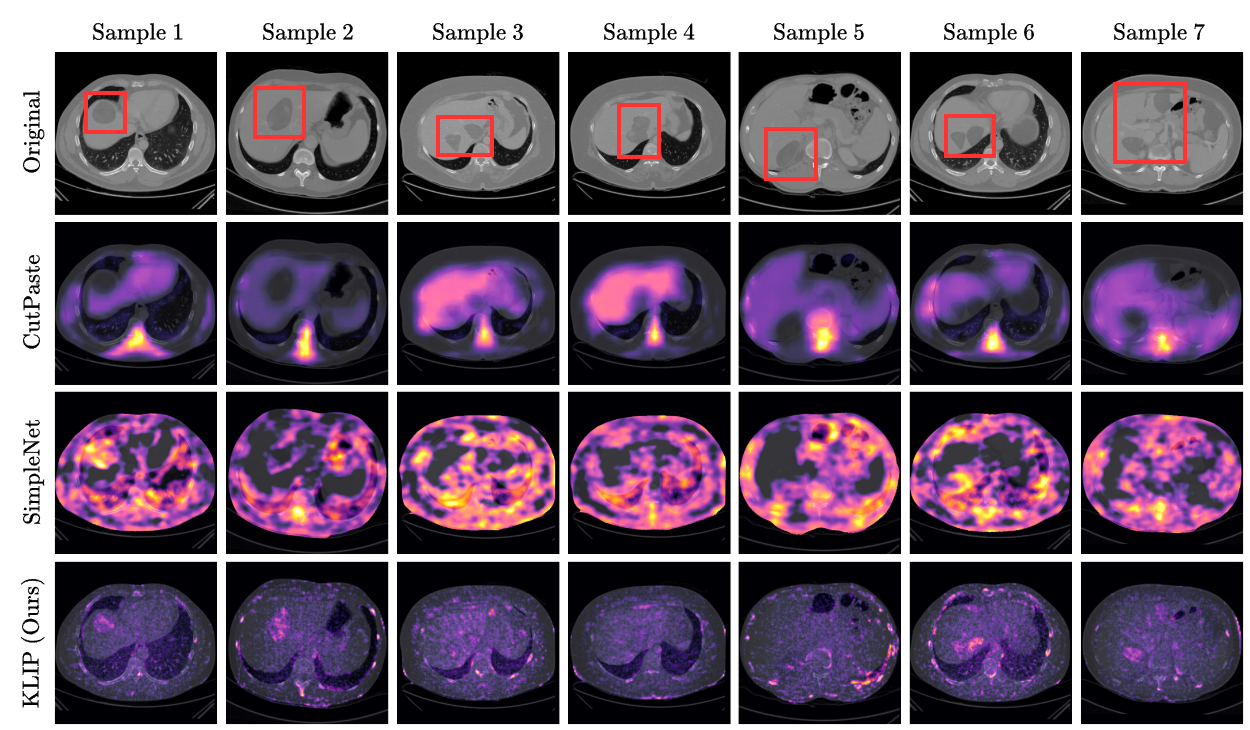}
    \vspace{-5mm}
    \caption{\textbf{Visual results for image-level OOD detection on sparse-view CT scans.} 
    \textit{Row 1:} Images in the OOD set with large tumors of medium darkness. Red boxes annotate where the tumors are.
    \textit{Rows 2-4:} Heatmaps of CutPaste \cite{cutpaste}, SimpleNet \cite{simplenet}, and KLIP overlaid on images.}
    \vspace{-4mm}
   \label{fig:sup_tumor_medium_large}
\end{figure*}

\begin{figure*}[ht]
  \centering
    \includegraphics[width=1.0\textwidth]{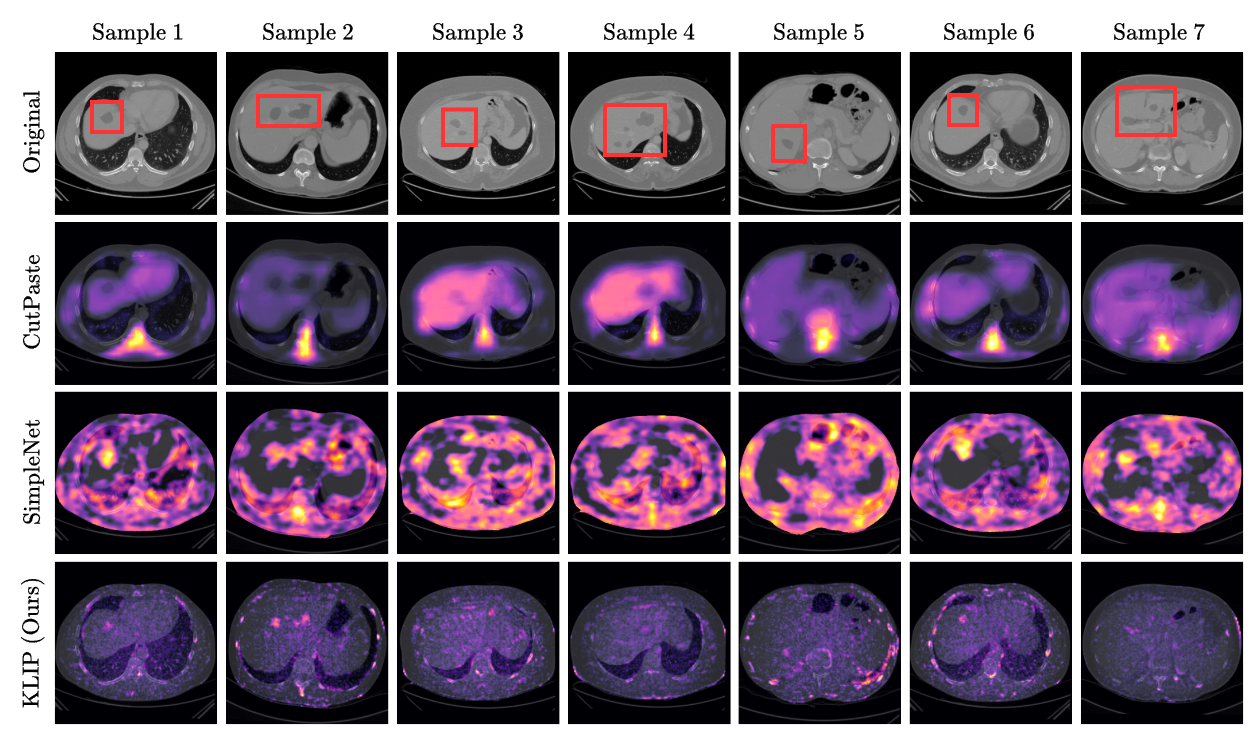}
    \vspace{-5mm}
    \caption{\textbf{Visual results for image-level OOD detection on sparse-view CT scans.} 
    \textit{Row 1:} Images in the OOD set with small tumors of medium darkness. Red boxes annotate where the tumors are.
    \textit{Rows 2-4:} Heatmaps of CutPaste \cite{cutpaste}, SimpleNet \cite{simplenet}, and KLIP overlaid on images.}
    \vspace{-4mm}
   \label{fig:sup_tumor_medium_small}
\end{figure*}

\begin{figure*}[ht]
  \centering
    \includegraphics[width=1.0\textwidth]{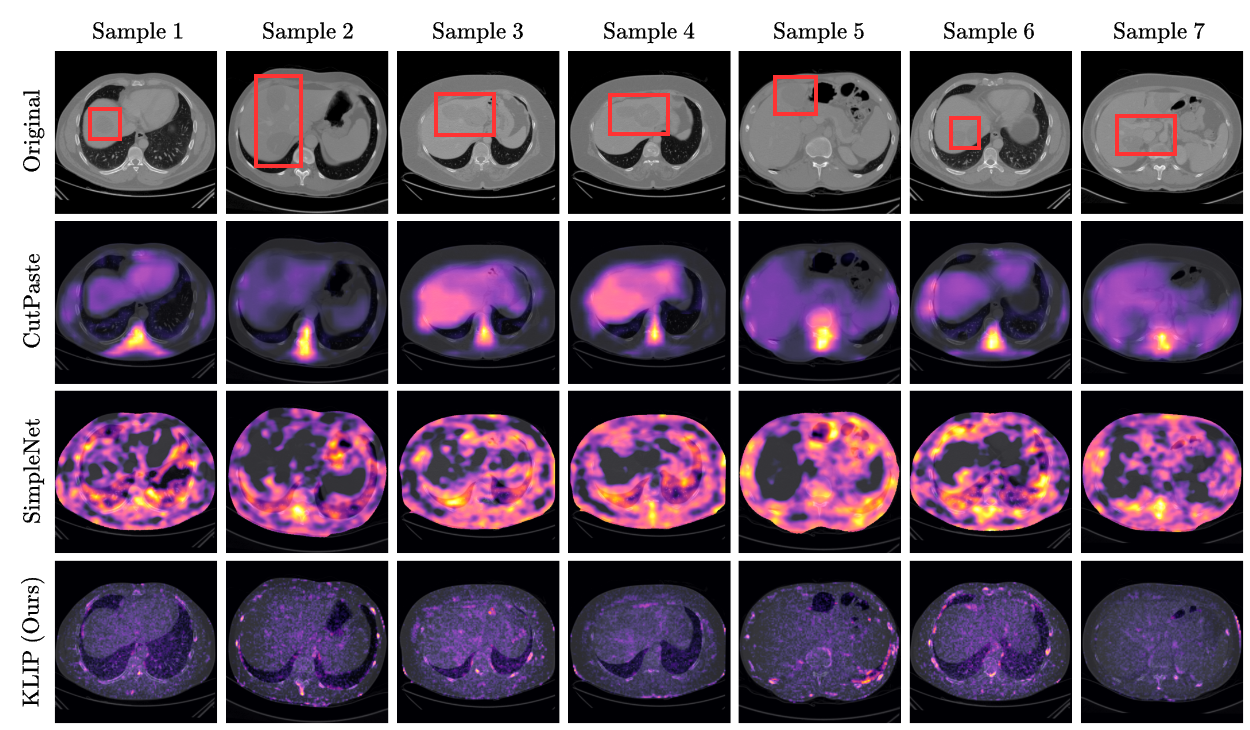}
    \vspace{-5mm}
    \caption{\textbf{Visual results for image-level OOD detection on sparse-view CT scans.} 
    \textit{Row 1:} Images in the OOD set with light and large tumors. Red boxes annotate where the tumors are.
    \textit{Rows 2-4:} Heatmaps of CutPaste \cite{cutpaste}, SimpleNet \cite{simplenet}, and KLIP overlaid on images.}
    \vspace{-4mm}
   \label{fig:sup_tumor_light_large}
\end{figure*}

\begin{figure*}[ht]
  \centering
    \includegraphics[width=1.0\textwidth]{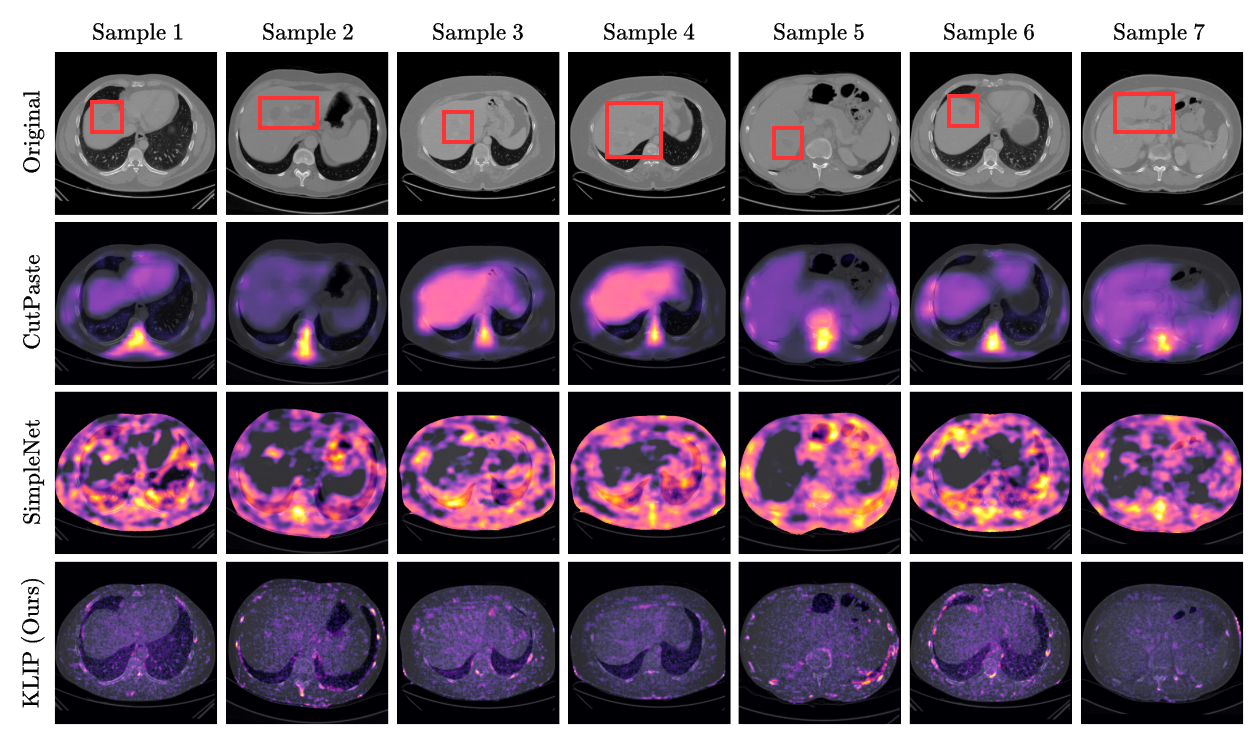}
    \vspace{-5mm}
    \caption{\textbf{Visual results for image-level OOD detection on sparse-view CT scans.} 
    \textit{Row 1:} Images in the OOD set with light and small tumors. Red boxes annotate where the tumors are.
    \textit{Rows 2-4:} Heatmaps of CutPaste \cite{cutpaste}, SimpleNet \cite{simplenet}, and KLIP overlaid on images.}
    \vspace{-4mm}
   \label{fig:sup_tumor_light_small}
\end{figure*}

For image-level OOD detection, \name{} shows strong performance when applied to both the predictor-corrector based diffusion model and the patch-based diffusion model. 
In both cases, AUC metrics degrade gradually with decreasing tumor size and darkness. 
\Cref{fig:sup_star} to \Cref{fig:sup_tumor_dark_large} visually compare the localization of baseline metrics and \name{}.
\Cref{fig:sup_star} shows examples of images with star artifacts, while the figures from \Cref{fig:sup_tumor_dark_large} to \Cref{fig:sup_tumor_light_small} differ only in the size and darkness of the simulated tumors.
All figures follow the same structure: the first row contains the sampled images, the second row shows the CutPaste \cite{cutpaste} heatmap overlays, the third row presents the SimpleNet \cite{simplenet} heatmap overlays, and the fourth row displays the results of our method \name{} using the predictor–corrector diffusion model \cite{song2021solving}.
While the baseline methods accurately detect the star artifacts, they struggle to locate tumors of any size or darkness, instead highlighting other regions of the anatomy that are actually in-distribution. Although \name{} does not have perfect tumor detection either, it shows much stronger tumor localization than either baseline, and is especially adept at detecting darker tumors.

\section{Forward Model Mismatch}
\label{sec:modelmismatch}

\begin{figure}
    \centering
    \includegraphics[width=1\linewidth]{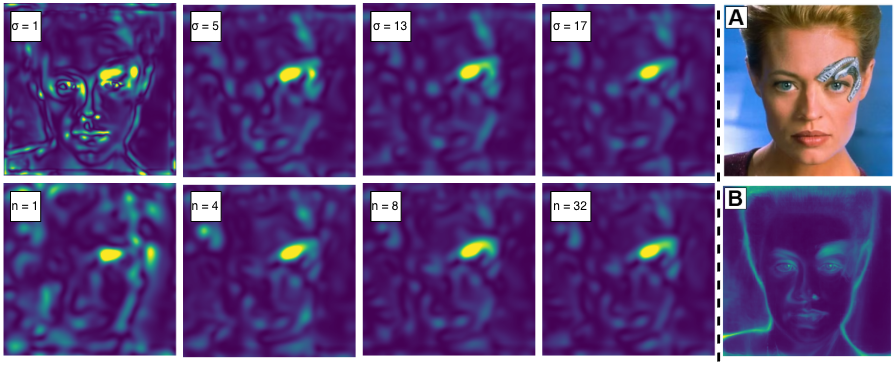}
    \caption{
    Left of the dotted line: KLIP heatmaps. 
    Top: forward model mismatch---measurements use a $21\times21$ Gaussian blur kernel with $\sigma=9$, while reconstruction uses kernels with $\sigma\in\{1,5,13,17\}$ (KLIP computed with 8 samples). 
    Bottom: effect of the number of samples $\{1,4,8,32\}$ with no model mismatch. 
    Right of the dotted line: (A) original image; (B) pixel-wise 95\% conformal CI length for the sample mean over 128 samples.
}
    \label{fig:different_kernel}
\end{figure}

Following prior work (e.g.,~\cite{song2021solving,dps}), our main experiments assume a matched forward model, meaning that the forward model in the measurement and reconstruction are same. This setting allows us to evaluate \name{} without the additional effect of forward model misspecification. However, exact knowledge of the forward model may be unavailable in practice, and reconstruction often relies on only an approximate operator. To study the sensitivity of \name{} to model mismatch, we evaluate \name{} in the Gaussian deblurring task by varying the reconstruction blur kernel while keeping the true measurement blur kernel fixed. As shown in \Cref{fig:different_kernel} and \Cref{tab:kernel_mismatch}, \name{} remains effective under such mismatch, although its precision gradually degrades as the assumed blur kernel moves away from the true one. 

\begin{table}[t]
\centering
\caption{Average image-level AUC of \name{} under reconstruction-model mismatch on 100 CelebA images with synthetic scar artifacts. The score is computed using 8 samples. The true measurement blur kernel is fixed at $\sigma=9$, while the reconstruction blur kernel is varied. Performance is best under the matched setting and degrades gracefully under moderate mismatch.}
\label{tab:kernel_mismatch}
\begin{tabular}{c c c c c c}
\toprule
$\sigma_{\mathrm{rec}}$ & 3 & 7 & 9 & 11 & 13 \\
\midrule
$\mathrm{AUC}_{\mathrm{avg}}$ & 0.809 & 0.897 & 0.904 & 0.898 & 0.893 \\
\bottomrule
\end{tabular}
\end{table}

\section{Computational Evaluation}
\label{sec:computationalcost}
We also compare the computational cost of \name{} against baseline OOD detection methods in \Cref{tab:runtime}. We report the time to sample 8 reconstructions at 512$\times$512 from a single CT measurement $y$ using ~\cite{song2021solving}, and the total runtime of each method. 
\name{} takes only $\sim$ 2\% longer than sampling alone. Results indicates that \name{} achieves competitive detection performance while remaining computationally close to the underlying reconstruction pipeline.

\begin{table}[t]
\vspace{-1em}
    \renewcommand{\arraystretch}{0.7}
    \centering
    \caption{\textbf{Runtime comparison} in seconds. ``Sampling'' denotes the time required to generate 8 reconstructions at $512\times512$ from a single CT measurement $y$. The remaining entries report the total runtime of each method, including reconstruction and scoring.}

    \label{tab:runtime}
    \begin{tabular}{ccccccc}
        \toprule
                \emph{\footnotesize	 Sampling}
                & \footnotesize	 NLL
                & \footnotesize	 DiffPath
                & \footnotesize	 SimpleNet
                & \footnotesize	 Cutpaste
                & \footnotesize	 KLIP
                \\
        \midrule
                \footnotesize	375
                & \footnotesize	518
                & \footnotesize	548
                & \footnotesize	376
                & \footnotesize	376
                & \footnotesize	383
                \\
        \bottomrule
        \vspace{-8mm}
    \end{tabular}
\end{table}

\section{Sample Size Sensitivity}
\label{sec:sampling}

Since \name{} estimates an expectation through Monte Carlo sampling, we evaluate how sensitive its performance is to the number of samples used in approximating the expectation. As shown in \Cref{fig:different_kernel} and \Cref{tab:sample_sensitivity}, \name{} is reasonably stable across different sampling budgets, with performance improving as more samples are used, but with diminishing returns beyond a moderate number of samples. In particular, the average image-level AUC over 100 CelebA images with synthetic scar artifacts increases from $0.813$ with a single sample to $0.904$ with 8 samples, and then changes only marginally for larger sample sizes. These results suggest that a moderate sampling budget already provides a reliable approximation, offering a favorable trade-off between computational cost and OOD detection performance.

\begin{table}[t]
\centering
\caption{Average image-level AUC of \name{} on 100 CelebA images with synthetic artifacts for different Monte Carlo sample sizes used to approximate the expectation in \name{}. Here, $N$ denotes the number of samples, and AUC denotes the average image-level OOD detection performance. Performance improves with larger $N$, but the gain becomes marginal beyond 8 samples.}
\label{tab:sample_sensitivity}
\begin{tabular}{c c c c c c}
\toprule
$N$ & 1 & 4 & 8 & 16 & 32 \\
\midrule
AUC & 0.813 & 0.854 & 0.904 & 0.905 & 0.909 \\
\bottomrule
\end{tabular}
\end{table}

\begin{table*}[ht]
    \centering
    \caption{\textbf{AUC results for dataset-level and image-level OOD detection, for different OOD artifacts and models on a sparse-view CT inverse problem.} We mark with $\dagger$ those AUC values that correspond to settings for which we tuned the \name{} hyperparameters. Other experiments use the same set of hyperparameters without further refinement. The best AUC for each task and dataset is underlined.}
    \vspace{-2mm}
    \label{tab:sup_ablation}
    \begin{tabular}{llccccccc}
        \toprule
            &
            & \multicolumn{6}{c}{\textbf{Tumor}}
            & \multirow{3}{*}{\textbf{Star}}
            \\
        \cmidrule(lr){3-8}
            &
            & \multicolumn{2}{c}{\textbf{Light}}
            & \multicolumn{2}{c}{\textbf{Medium}}
            & \multicolumn{2}{c}{\textbf{Dark}}
            &
            \\
        \cmidrule(lr){3-4} \cmidrule(lr){5-6} \cmidrule(lr){7-8} 
            &
            & Small
            & Large
            & Small
            & Large
            & Small
            & Large
            & 
            \\
        \cmidrule(lr){1-9}
            \multirow{12}{*}{\rotatebox[origin=c]{90}{\textbf{Dataset Level}}}
            & CutPaste \cite{cutpaste}
            & 0.491 & 0.485 & 0.487 & 0.505 & 0.484 & 0.506 & \underline{0.999}
            \\
            &
            SimpleNet \cite{simplenet}
            & 0.499 & 0.501 & 0.505 & 0.504 & 0.499 & 0.501 & 0.993
            \\
            &
            Predictor-Corrector \cite{song2021solving}
            & & & & & & &
            \\
            &
            \hspace{1.5em}NLL
            & 0.511 & 0.514 & 0.528 & 0.535 & 0.540 & 0.559 & 0.586
            \\
            &
            \hspace{1.5em}DiffPath
            & 0.334 & 0.342 & 0.349 & 0.368 & 0.370 & 0.409 & 0.688
            \\
            &
            \hspace{1.5em}$D_{KL}$
            & 0.531 & 0.537 & 0.580 & 0.602 & 0.600 & 0.621 & 0.541
            \\
            &
            \hspace{1.5em}\name{} (Ours)
            & \underline{0.754} & \underline{0.774} & \underline{0.772} & \underline{0.776} & \underline{0.772} & \underline{0.782} & 0.855$^\dagger$
            \\
            &
            PaDIS \cite{padis}
            & & & & & & &
            \\
            &
            \hspace{1.5em}NLL
            & 0.498 & 0.478 & 0.525 & 0.490 & 0.469 & 0.460 & 0.502
            \\
            &
            \hspace{1.5em}DiffPath
            & 0.469 & 0.482 & 0.507 & 0.497 & 0.504 & 0.481 & 0.480
            \\
            &
            \hspace{1.5em}$D_{KL}$
            & 0.498 & 0.534 & 0.510 & 0.545 & 0.538 & 0.643 & 0.506
            \\
            &
            \hspace{1.5em}\name{} (Ours)
            & 0.499 & 0.513 & 0.510 & 0.502 & 0.523 & 0.512 & 0.512
            \\
        \cmidrule(lr){1-9}
            \multirow{8}{*}{\rotatebox[origin=c]{90}{\textbf{Image Level}}}
            & CutPaste \cite{cutpaste}
            & 0.754 & 0.695 & 0.651 & 0.441 & 0.651 & 0.319 & 0.978
            \\
            &
            SimpleNet \cite{simplenet}
            & 0.378 & 0.216 & 0.822 & 0.592 & 0.920 & 0.657 & \underline{0.999}
            \\
            &
            Predictor-Corrector \cite{song2021solving}
            & & & & & & &
            \\
            &
            \hspace{1.5em}$D_{KL}$
            & \underline{0.799} & \underline{0.800} & \underline{0.890} & 0.856 & 0.906 & 0.853 & 0.837
            \\
            &
            \hspace{1.5em}\name{} (Ours)
            & 0.785 & 0.791 & 0.876 & \underline{0.878} & 0.920 & 0.904 & 0.912$^\dagger$
            \\
            &
            PaDIS \cite{padis}
            & & & & & & &
            \\
            &
            \hspace{1.5em}$D_{KL}$
            & 0.595 & 0.663 & 0.857 & 0.672 & \underline{0.932} & \underline{0.941} & 0.841
            \\
            &
            \hspace{1.5em}\name{} (Ours)
            & 0.667 & 0.630 & 0.859 & 0.732 & 0.911 & 0.852 & 0.889
            \\
        \bottomrule
    \end{tabular}
\end{table*}

\end{document}